\documentclass[twocolumn]{article}          

\usepackage{graphicx}
\usepackage{epstopdf}
\usepackage{cprotect}
\usepackage{ marvosym }

\usepackage{algorithmic}
\usepackage{algorithm}
\usepackage{listings}
\usepackage{textcomp}
\usepackage[table]{xcolor}

\usepackage[american]{babel} 

\definecolor{darkred}{rgb}{0.5,0.0,0.0}
\definecolor{lightgray}{rgb}{0.9,0.9,0.9}
\definecolor{lightblue}{rgb}{0.3,0.3,1}
\definecolor{lightgreen}{rgb}{0.2,0.6,0.2}
\definecolor{darkyellow}{rgb}{0.6,0.6,0.0}

\usepackage{times}
\usepackage{epsfig}
\usepackage{epstopdf}
\usepackage{graphicx}
\usepackage{amsmath}
\usepackage{amssymb}

\usepackage{multirow}

\usepackage{color}
\usepackage{subfigure}
\usepackage{multirow}

\usepackage{mathtools}
\usepackage{url}

 \usepackage{amsfonts}
\usepackage{amssymb}
 \usepackage{theorem}
\usepackage{trfsigns}
\usepackage{subfigure}
\include {multirow}

\usepackage{makecell}
\usepackage{tabularx}

\newcommand{\Cc}{\mathbb C}

\newcommand{\ji}{j}

\usepackage[margin=2.5cm]{geometry}

\newtheorem{defi}{\textbf{Definition}}

\newcommand{\Complex}[0]{\mathbb{C}}
\newcommand{\Real}[0]{\mathbb{R}}
\newcommand{\Natural}[0]{\mathbb{N}}

\newcommand{\T}[0]{\top}
\newcommand{\gmv}[1]{\mbox{\boldmath$#1$}}
\newcommand{\mv}[1]{\mathbf{#1}}

\newcommand{\PA}[3]{#2\circ_{\!#1}#3}

\usepackage{authblk}

\author[1,2,3]{Marco Reisert}
\author[1,3]{Volker A. Coenen}
\author[1]{Christoph Kaller}
\author[1,4]{Karl Egger}
\author[5]{Henrik Skibbe}
\affil[1]{Medical Center, Freiburg University, Germany}
\affil[2]{Department of Medical Physics}
\affil[3]{Department of Stereotactic and Functional Neurosurgery}
\affil[4]{Department of Neuroradiology}
\affil[5]{Ishii-Lab, Department of Systems Science, Graduate School of Informatics, Kyoto University, Japan}

\begin{document}

\selectlanguage{american}

\title{HAMLET: Hierarchical Harmonic Filters for Learning Tracts from Diffusion MRI }

\maketitle

\begin{abstract}

In this work we propose HAMLET, a novel tract learning algorithm, which, after training, maps raw diffusion weighted MRI directly onto an image which simultaneously indicates tract direction and tract presence. 
The automatic learning of fiber tracts based on diffusion MRI data is a rather new idea, which tries to overcome limitations of atlas-based techniques.
HAMLET takes a such an approach. Unlike the current trend in machine learning, HAMLET has only a small number of free parameters
HAMLET is based on spherical tensor algebra which allows a translation and rotation covariant treatment of the problem. HAMLET is based on a repeated application of convolutions and non-linearities, which all respect the rotation covariance. The intrinsic treatment of such basic image transformations in HAMLET allows the training and generalization of the algorithm without any additional data augmentation. We demonstrate the performance of our approach for twelve prominent bundles, and show that the obtained tract estimates are robust and reliable.  It is also shown that the learned models are portable from one sequence to another.

\end{abstract}

\section{Introduction}

It becomes more and more apparent that looking through the eyes of diffusion MRI (dMRI)
shows rather  a rough sketch of the underlying connectome. It was shown that tracking algorithms 
running on brain-like numerical phantoms with known ground truth heavily 
suffer from false positives \cite{maier2017challenge}.
There have been attempts to account for such problems
by reweighing and filtering strategies. Of course, also data quality plays a role, low
spatial and angular resolution as experienced in the clinics is not sufficient and 
introduces ambiguities that cannot be resolved with pure model-based approaches.
Explicit usage of anatomical prior knowledge is necessary. Atlas-guided methods \cite{hagler2009automated} do not 
suffer from these problems, because the existence of a fiber bundle is taken as 
a prerequisite and the position in atlas-space is very informative.
However, those methods rely on an accurate registration with an 
anatomical contrast. Also the inclusion of the tract prior knowledge is usually on a
local level, i.e. the progression probabilities during streamlining are multiplied
by priors defined locally. In this work we follow an alternative approach and
use machine learning. We will directly map the dMRI data onto 
a directional saliency map of tract presence. It is nearby to require such 
a mapping to be translation covariant, as convolution neural networks (CNN) are used to 
be. We go here a step further and require additionally the filter
to be rotation covariant. Which is, if we fully refrain from using a template
space or normalization frame, also a very reasonable assumption. The idea
we propose here for representation of such a mapping, we call it rotation covariant filters,
is based on Spherical Tensor Algebra \cite{skibbespherical}. The filter is, similar to CNNs, an iterated
application of spatial convolutions and non-linearities, however, all operations
of the filter are rotation covariant, i.e. wherever one rotates the input data, 
the output is rotated accordingly. So, the filter will only learn the typical 
spatio-orientational relationships between white matter structures and 
nothing about their absolute orientations nor positions. It has the advantage
that the filter does not have to learn the robustness against
orientational changes from the training data, but has it intrinsically built in. The
filter can generalize quick and does not need any data augmentation.

The outputs of HAMLET are spherical tensor fields of order $2$, or, equivalently  fields of trace-less, symmetric $3\times3$ matrices. The magnitude of the tensors
represent the local "likeliness" for the presence of a specific tract, the tensor orientation predicts the local tract directions. 
The top of Figure \ref{fig:overview} shows an overview of the principle. Once the tensor field is predicted several interpretations are possible.
It is nearby to use the field for a subsequent streamline tracking, which is also the way we follow mostly in this study.

There are two main technical contribution in this paper: (1) In HAMLET, we use an extension of the scalar filter proposed in \cite{skibbespherical}. Unlike its scalar valued predecessor, the extended filter in HAMLET can now learn and output directions in addition to a tract saliency score. (2) In order to determine the tract locations and orientations, a global treatment of the problem is proposed. The large amount of 3D image data requires a hierarchical, multi-scale 
implementation to incorporate global image features for the rough tract localization, and local features for the precise representation of fiber bundle properties on a voxel resolution level.

Practically, we provide a fully automatic tracking and tract labeling  procedure for (known)
anatomical white matter fiber bundles. We will show that the procedure can measure
apparent tract volumes in a reliable way and provides nice visualizations 
and representations of white matter geometry, even in a clinical setting.
In its current implementation the filter relies only on second-order orientation information of the dMRI data, and is hence applicable to large varieties of protocols. We will show that a filter can also work for a dMRI protocol it was not trained with.

\subsection{Related Work}
Machine learning in the context of diffusion MRI was already used for simple white/gray matter segmentations \cite{schnell2009fully},
or learning brain parcellations/segmentations \cite{skibbe2011gaussian,skibbe2011dense}.
Also in the context of microstructure estimation machine learning approaches provide 
efficient estimates \cite{reisert2017disentangling}. 
In \cite{neher2015machine,jorgens2018learning} neural networks are used to
predict progression probabilities for
streamline tracking. In \cite{poulin2017learn} these ideas are extended to tract shape. 
Also the direct learning of fiber orientation distributions is possible (\cite{koppers2016direct,koppers2017reconstruction}).

The very recent work TractSeg of \cite{wasserthal2017direct,tractseg} is closest to the spirit of what we propose here,
however lacks of intrinsic invariance properties, and thus, requires heavy data augmentations.

\begin{figure*}[t]
\includegraphics[width =16cm]{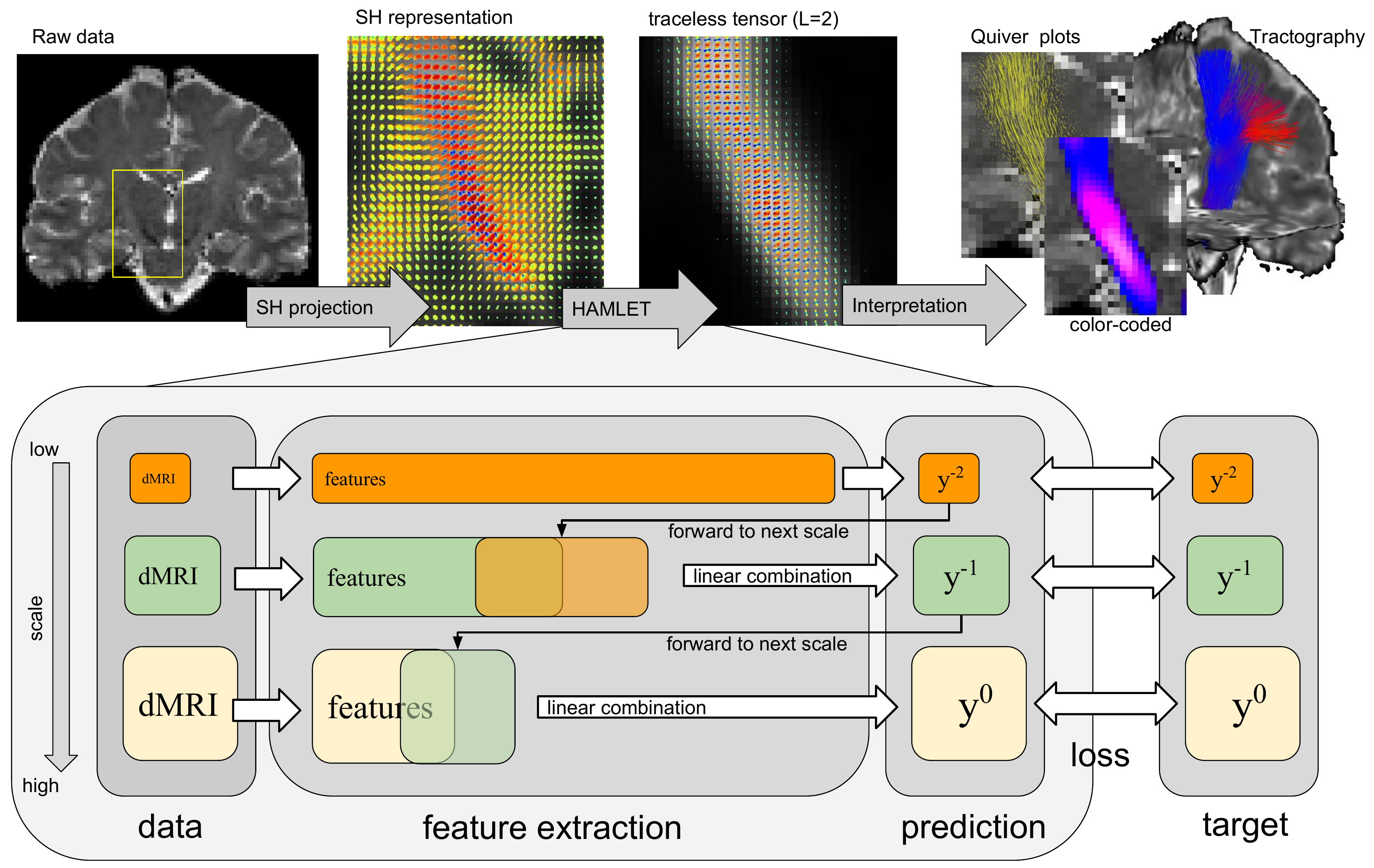}
\caption{The rough workflow of the algorithm: the input to HAMLET are L=0,2 spherical harmonic projections on different scales.}
\end{figure*}


\section{Methods}
In the following we give a brief introduction to spherical tensor algebra and show 
how it can be used to build rotation covariant trainable mappings. For a more complete introduction see \cite{skibbespherical}.
\subsection{Spherical Tensor Algebra}
Spherical harmonic representations are quite commonly used in the dMRI community to represent the direction dependent diffusion signal.

Let $x:\Real^3\times S_2 \mapsto \Complex$ be a function which varies with image coordinates $\mv r\in \Real^3$ and directions $\mv n\in S_2$. A nearby example is a dMRI signal. We call such a function an orientation field. It should be noted that we consider, without any loss of generality, that $x$ is complex valued. The field $x$ can be expressed in terms of the spherical harmonic expansion 
\begin{align}
x(\mv r, \mv n) = \sum_{j=0}^\infty \mv a^j(\mv r)^\T \mv Y^j({\mv n}). 
\label{eq:orientationfield}
\end{align}
The functions $\mv Y^j: S_2 \mapsto \Complex^{2j+1}$ are the orthogonal spherical harmonic basis functions of order $j$. The expansion coefficients $\mv a^j(\mv r)\in  \Complex^{2j+1}$ form themselves a field  $\mv a^j : \Real^3 \mapsto \Complex^{2j+1}$. We call such a field of spherical harmonic coefficients a spherical tensor field of order $j$ (ST-field).  

Any orientation field $x$ can be rotated with $(gx)(\mv r,\mv n) := x(\mv R(g)^\T\mv r,\mv R(g)^\T\mv n)$, where the first argument is a coordinate transformation, and the second argument a value transformation that rotates the local spherical function accordingly. For the corresponding expansion fields this means,
\begin{align}
(gx) (\mv r, \mv n) =  \sum_{j=0}^\infty (\mv D^\ji(g) \mv a^j(\mv R(g)^\T \mv r))^\T \mv Y^j({\mv n}),
\end{align}
where $\mv D^\ji(g)$ is the Wigner-D matrix, the irreducible representations of $SO(3)$.
This fact can be used as the defining one:
\begin{defi}[Spherical Tensor Field]
A function $\mv x^\ji : \Real^3 \mapsto \Complex^{2j+1}$ is called a spherical tensor field of order $j$ if it transforms
with respect to rotations as
\begin{align}
(g \mv x^\ji)(\mv r) := \mv D^\ji(g) \mv x^\ji( \mv R(g)^\T \mv r)
\label{eq::stensor::rot}
\end{align}
for all $g\in SO(3)$. 
\end{defi}

Spherical Tensor Algebra defines a set of rotation covariant operations on such fields. The most important operation is the spherical product:
\begin{defi} [Spherical Products]
For every $j \geq 0$ we define a family of bilinear forms, a tensor product
\begin{align}
&&\PA{j}{}{} : \Complex^{2j_1+1} \times \Complex^{2j_2+1} \mapsto \Complex^{2j+1}~,
\end{align}
where $j_1,j_2 \in \Natural$ has to be chosen according to the triangle inequality 
$|j_1 - j_2| \leq j \leq j_1 + j_2$. The product is defined by
\[
(\mv e^j_m)^\T (\PA{j}{\mv v}{\mv w}) :=\hspace{-0.5cm} \sum_{m = m_1+m_2}\hspace{-0.2cm} \langle j m \mid j_1 m_1, j_2 m_2 \rangle
v_{m_1} w_{m_2}~.
\]
where $\mv e^j_m$ is the $m$th standard basis vector in $\Cc^{2\ji+1}$ and $\langle j m \mid j_1 m_1, j_2 m_2 \rangle$ are the Clebsch-Gordan coefficients.
\end{defi}
Spherical Products are covariant in the sense that 
$\PA{j}{(\mv D^{j_1}(g)\mv v)}{(\mv D^{j_2}(g)\mv w)} = \mv D^{j}(g) (\PA{j}{\mv v}{\mv w})$
for any $\mv v \in \Complex^{2j_1+1}$ and $\mv w \in \Complex^{2j_2+1}$, and any $g\in SO(3)$.

On the basis of spherical products spherical derivatives can be introduced.
The homogeneous polynomials $R^j_m(\mv r) = r^j Y^j_m(\mv n)$ of order $j$ are called the ``solid harmonics``; 
With  $r = |\mv r|$ we denote the distance to the center, and with $\mv n = \mv r/r$ the direction. 
The operator transformation $R^j_m(\nabla)$ maps the Cartesian gradient operator $\nabla=(\partial_x,\partial_y,\partial_z)^T$ onto the spherical domain and one 
can define a spherical differential operator $\gmv \partial^\ji:=(\partial^\ji_{-\ji}, \cdots, \partial^\ji_{\ji})^T$ by 
      \begin{align}
	  \partial^j_m :=R^j_m(\nabla).
      \end{align}     
The operator can be used to linearly map ST-fields onto other ST-fields by intertwining spatial neighborhood information. 
From a computational perspective this derivative operator can be used to efficiently compute local projection on spherical basis
function.
  
\subsection{Covariant Nonlinear Filters}
The basic concept we rely on was already presented here \cite{} for scalar-valued output (spherical order $j=0$). 
As we are interested in learning images of tracts we want to extend this here to second-order (spherical order $j=2$) tensor-like output. But first, the general concept. Suppose we 
have given a collection of spherical tensor fields $D = \{ \mv a^{0}, \mv a^{2}, \hdots\mv a^{L_\text{max}} \}$ describing the dMRI data, i.e. the SH-projections of 
a dMRI signal at a certain b-value. Note that in all experiments we set $L_\text{max}=2$,
i.e. already a rather low number of gradient directions will be sufficient.
We can now form linear features describing local neighborhoods by the following differential operations:
\[
\mv b_{J,L,j,g} = \PA{J}{\gmv \partial^L}{(g * \mv a^{j})},
\]
where $g*$ is a convolution with a radial function, typically a Gaussian. Due to
the commutativity of convolutions the above operation may be interpreted as a projection
onto  the by $\PA{J}{\gmv \partial^L}{g}$ generated basis functions. For a Gaussian $g$,
they are related to the Gauss-Laguerre basis.
For simplicity we set here $I = (J,L,j,g)$ including all free indices/parameters on the right hand side. Note the constraints on the indices
coming from the selection rules of the Clebsch-Gordan coefficients, hence, not all combinations are possible. The linear features $\mv b_I$ can 
now be used to form non-linear ones by spherical products 
$\PA{j}{\mv b_I}{\mv b_{I'}}$. One can form also higher order products than order two, but here
we refrain from doing so. Indeed, the predicted tract labels
are homogeneous of order two in the b0-normalized dMRI signal. 
Now, these products can again be linearly mapped by differentiation 
to any possible order. As we want to learn tracts, we map them to order two:
\[
\mv f_{j}(\mv b_I,\mv b_{I'})= \PA{2}{\gmv \partial^L }{(g * (\PA{j}{\mv b_I}{\mv b_{I'}}) )}
\]
Finally, linear combinations 
of such  $\mv f$ form the final filter output, i.e. the input $D$ containing the 
dMRI data is mapped onto 
\[
\mv y(D) = \sum_{j,I,I'} \alpha_{jII'}\ \mv f_{j}(\mv b_I,\mv b_{I'})
\]
where $\alpha_K$ are coefficients to be determined during a training stage. 
Due to the linearity in $\alpha_K$ the training can easily be accomplished in 
a least-squares sense by solving a linear equation. The number of features, i.e.
the number of combinations for the multi-index $K$ can take, is crucial, it can easily get
quite high, we denote it by $N$. 
%

\subsection{Hierarchical Covariant Filters}
The described approach works already quite well for simple tasks, however,  
problems appear when the learning task gets difficult. Increasing the complexity of the 
filter by increasing the number of features $\mv f$ by allowing more product
combinations or higher orders is possible, however, one can easily get into 
trouble regarding the memory consumption and computation time. One dataset can
easily take tenth of Gigabytes of memory ($N \approx 1000$). We propose here a simple hierarchical multi-scale 
generalization of the above filter to increase the expressiveness of the filter
while keeping computation and memory requirement in a reasonable range. The filter
is consecutively applied: it starts at low resolution with a rather high number 
of features $N$ and the output is forwarded to the next higher scale.
To be precise: let $\mv y^s$  denote the result of the filter at scale $s$ and $\mv b_I^{s}$ the linear features from equation \ref{eq} computed on properly rescaled versions of the data at scale s,
then the filter output is defined by
\begin{eqnarray}
\mv y^{s+1} &:=& \kappa^s \mv y^{s} + |\mv y^{s}|  \sum_ {j,I,I'} \alpha^s_{jII'}  \mv f_{j}(\mv b_I^{s+1},\mv b_{I'}^{s+1})\nonumber \\ 
&&+\  \sum_{j,I,k} \beta^s_{jIk}  \mv f_{j}(\mv b_I^{s+1}, \mv y_k^{s} ) \label{eq:feats}
\end{eqnarray}
where \[
\mv y_k^{s} := \PA{2k}{\mv y^s_{k-1}}{\mv y^s} / |\mv y_s|
\]
and $\mv y_1^s := \mv y^s$ with $k=1,2,3$.
The output is a linear combinations of: the output of the last scale, together with a quadratic term (as known from the single
scale filter) at the current scale windowed with the previous layer's magnitude response, and finally,
a term being a product of the signal with the normalized spherical powers of the previous output.

In all our experiments we used a rescaling factor of $\gamma=2$ and three scales $s=-2,-1,0$. 
The parameters $\kappa,\alpha,\beta$ are the factors to be learned. The number of features differ and 
decrease with increasing resolution to keep the memory consumption similar at each scale. Details how the 
the rescaling was done can be found in the appendix \ref{sec:learning}.

 \subsubsection{Parity and Point Symmetry}
 Besides rotation covariance the filter can also to be made to be invariant to parity changes
 (axial or point reflections in 3D),
 if the spherical products are chosen accordingly. It seems to be reasonable to require the filter 
 to be covariant with respect to mirroring at the saggital plane, which involves a parity change of 
 the coordinate system. In this way  the machine does not have any prejudices about hemispherical
 differences. Recall the features
 \[
\mv f_{(\ell,I,I')} = \PA{2}{\gmv \partial^L }{(g * (\PA{\ell}{\mv b_I}{\mv b_{I'}}))}
 \]
 For parity covariant filters the sum $\ell+J_I+J_{I'}+L$ has to be even.
 More details how the products are chosen can be found in the appendix \ref{sec:product}

 A consequence of the parity covariance is that the filter can intrinsically not decide
 whether a tract is in the right or left hemisphere, so it makes no sense to learn tracts
 individually on the left or right hemisphere. All training labels should contain both homologues
 tracts for intra-hemispheric connection, commissural tracts are already approximately 
 symmetric. So, only approximately parity symmetric training labels make sense. Note, that this 
 does not mean that the output image of the filter is point symmetric, the filter is covariant
 with respect to parity changes.

\subsection{Training}
Suppose, we have given a number of training tuples $(D_i,\mv y_i)$, were $D_i$ stands for the 
raw dMRI dataset of the $i$th subject and $\mv y_i$ for a properly generated spherical order 2 tract label map. Imagine the label map as an indicator map of the tract to be detected, with values $|\mv y|=1$ in regions where the tract is present and zero otherwise. The principal direction of the 
$\mv y$ should coincide with the direction of the tract.
For example, for a set of 'ground truth' tracts represented by their traces $\mv r_k$ with tangents $\mv n_k$
we can compute
\begin{equation}
\mv y (\mv r) = \sum_k \int \mv Y^2(\mv n(t))\ \delta(\mv r_k(t)-\mv r) dt \label{eq:genlm}
\end{equation}
and normalize $\mv y' = \mv y / (|\mv y| + \epsilon)$, where $\epsilon$ is small 
constant determining a threshold of detectability. Indeed, we could also use directly the
density map $\mv y$, however, then fiber density and detection probability would intertwine,
which is not wanted and difficult to control. 

For the training procedure we follow rather an ad-hoc strategy.
Training starts at the lowest resolution, with properly smoothed and downsampled 
data and label maps. Once lowest resolution is trained over the whole training set, the predictions 
of the trained filter at this resolution are upsampled and forwarded to create the upper level features. Then, training 
is performed on the this upper level, and so on, until finest resolution is reached.
Note, that with this approach we do not optimize for interdependencies of weights between the layers,
but learn the filter layer by layer. This approach should work well, if the intermediate predictions 
are reproducible and generalize well. In order to assure this, an
L2-regularization (Thikonov)
is necessary, which has to be stronger for low resolutions than for high resolutions. Although high regularizations 
degrade predictions on the training set, they help to generalize quicker and keep prediction more reliable though less accurate. 
For details of the regularization see the appendix \ref{sec:learning}.

\begin{figure*}
    \begin{center}
    \includegraphics[width=16cm]{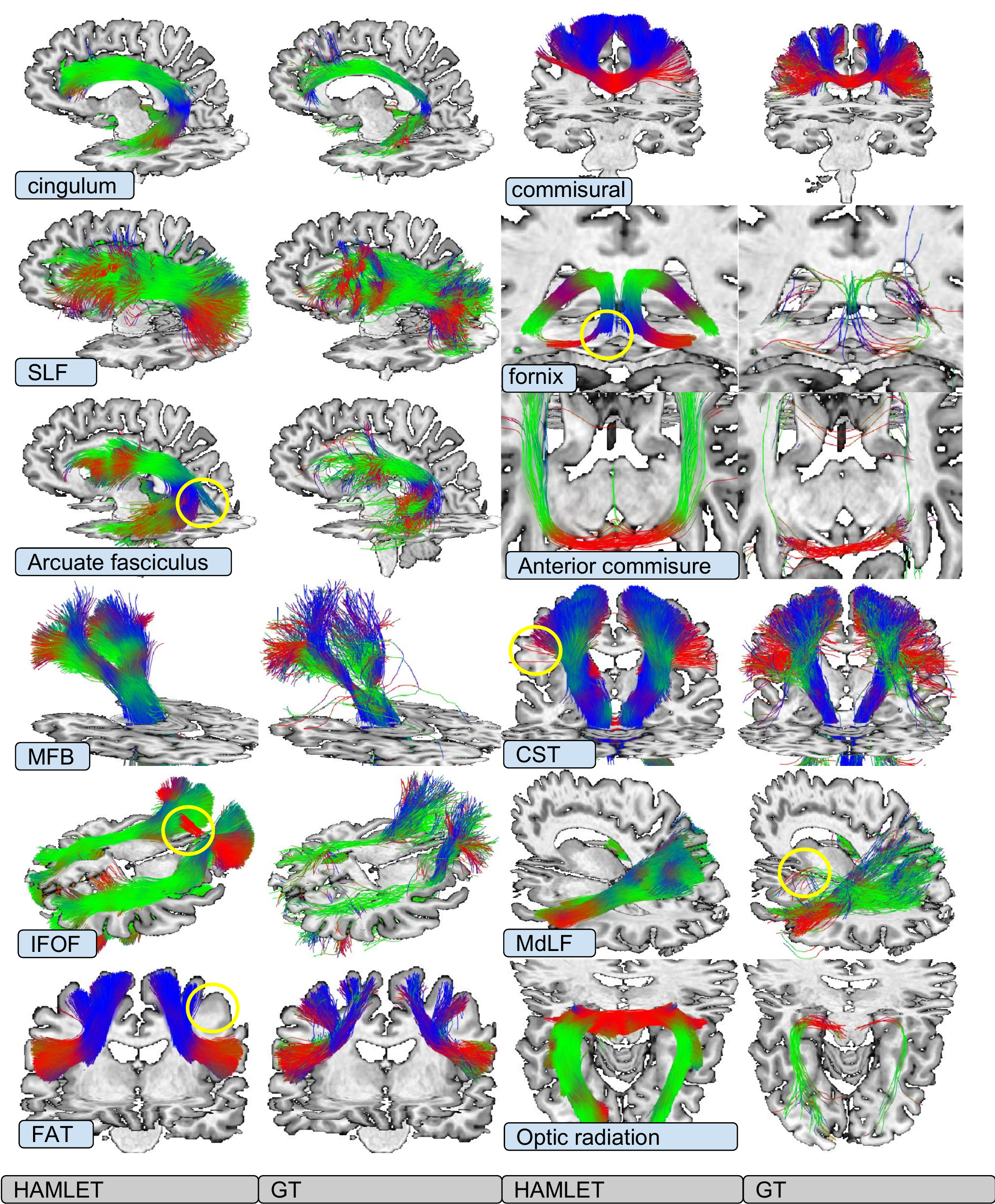}
    \end{center}
    \caption{Results of all considered tracts on a randomly chosen test subject.
    HAMLET is compared with results obtained from global tractography with automatic selection strategy. Issues (discussed in the text) are highlighted by yellow 
    circles.}
    \label{fig:overview}
\end{figure*}

\section{Data}
We consider two data sets, one rather high quality protocol measured at a Siemens
PRISMA for training and one low quality measured at a Tim TRIO. The first cohort
consists of 55 healthy controls (also used here \cite{coenen2018anatomy}), the second includes 28 
volunteers all scanned twice in different sessions for testing reliability.

(PRISMA) Healthy subjects were scanned on a Siemens 3 T TIM PRISMA using an SE EPI sequence with a TE = 88 ms and TR = 2008 ms, bandwidth 1780 Hz, flip-angle 90°, GRAPPA factor 2, SMS factor 3 with 17 non-diffusion weighted images, 2*58 images with b-factor b = 1000 and 2000 s/mm2; with an in-plane voxel size of 1.5 mm × 1.5 mm and a slice thickness of 3 mm. One phase-encoding flipped b = 0 image was acquired, which is used for distortion correction (FSL's top up (Andersson et al., 2003)). Additionally to the healthy subjects three tumor
patients will be shown to demonstrate how HAMLET works 
with unseen data.

(TRIO) Subjects were scanned on a  Siemens TIM TRIO using a 2-shell protocol with two shells at b-values 1 and  with 60 directions per shell, at an isotropic resolution of , 6/8 partial Fourier, TR=10,900 ms, TE=107 ms. The data was reconstructed with adaptive combine (Walsh et al., 2000) such that the noise distribution is close to Rician. Additionally, a T1-weighted structural dataset was acquired, resolution 1 mm isotropic. (maxim's dico)

\subsection{Preprocessing}
The diffusion weighted images were first denoised by a post-processing technique which uses random matrix theory (see \cite{veraart2016denoising} for details). This is followed by a Gibbs artifact removal 
\cite{kellner2016gibbs} based on local sub-voxel shifts. Then, images (from PRISMA) were corrected for EPI distortions by FSL's top up \cite{andersson2003correct}. The PRISMA dataset was additionally up-sampled to isotropic resolution by an edge preserving interpolation 
approach (Reisert et al., ISMRM 2017). Finally, on all scans underwent a simple
BET. For all subjects the anatomical reference scans were segmented using VBM8 (http://www.fil.ion.ucl.ac.uk/spm). White matter probability maps were thresholded at a probability of 0.5 to determine the area of fiber reconstruction. 

\subsection{Generation of Training Data}
To generate labels one may use any kind of classical fiber tracking algorithm at hand,
take a certain selection or seeding strategy to delineate the tract of interest and 
generate a tract images by equation \eqref{eq:genlm}. We decided to use global tractography
as described in \cite{reisert2010} and use point based selections to generate tract labels.
We selected the 12 tracts, which should cover most of the geometries typically
encountered: inter- and intra hemispherics, commissural and tracts originating or passing through the midbrain. Details about the selection criteria can be found in appendix \ref{sec:tract}. All selections are based on the non-rigid deformations to MNI group space provided by VBM8. We want to emphasize that the training set generation we used here is nothing than perfect. Still, false tracts can appear due to too loose restrictions, or tracts are missed because of slight atlas misregistrations.

\section{Results}

To get a first qualitative impression of the behavior 
Figure \ref{fig:overview} shows results from 
a random test subject from the PRISMA dataset. We compare with the automatic selected bundles 
from ordinary global tractography. The thresholding used for visualization was determined during training by maximization of segmentation accuracy.
The tracking itself was obtained from a simple Euler integration of the
vector field formed by the principal vector (eigenvector of the 
trace-less matrix corresponding to the maximal eigenvalue). Tracts are randomly
seeded within the volume and terminated at its boundary. No other 
stopping criteria are used.

\subsection{False Positives/Negatives}
The detected tracts follow mostly the automatically selected
GT tracts. However, also problems are apparent. 
Overall, the detections are rather 'smooth'. For example, 
for the MFB the bottleneck at its trunk is not accurately modeled. However, this is not always the case, for example
the fornix and the anterior commisure can be nicely detected,
although the structures are rather fine. However, 
the general problem exists that regions of a tract with 
high variability (over subjects) get lower responses in these regions.
So, depending on the threshold, the 'trunk' of a tract is 
always a bit enlarged, if also the fine tract processes 
should be included. For example, looking at the commisural 
fibers in the example in Figure \ref{fig:overview},
the laterals are not well represented with HAMLET in this 
subject. 

In general, false positive tracts are
small additional processes. For example, the AF shows sometimes a small additional extension at its posterior bend. Or the IFOF 
shows frontally a wrong inter-hemispheric connection. In fact, some
tracts are just simple to learn and others (e.g. the optic radiation) 
are more difficult. One can also find false positives in
regions with high fractional anisotropy. 

\subsection{Retest Reliability}
\begin{figure*}[t]
    \begin{center}
    \includegraphics[width=16cm]{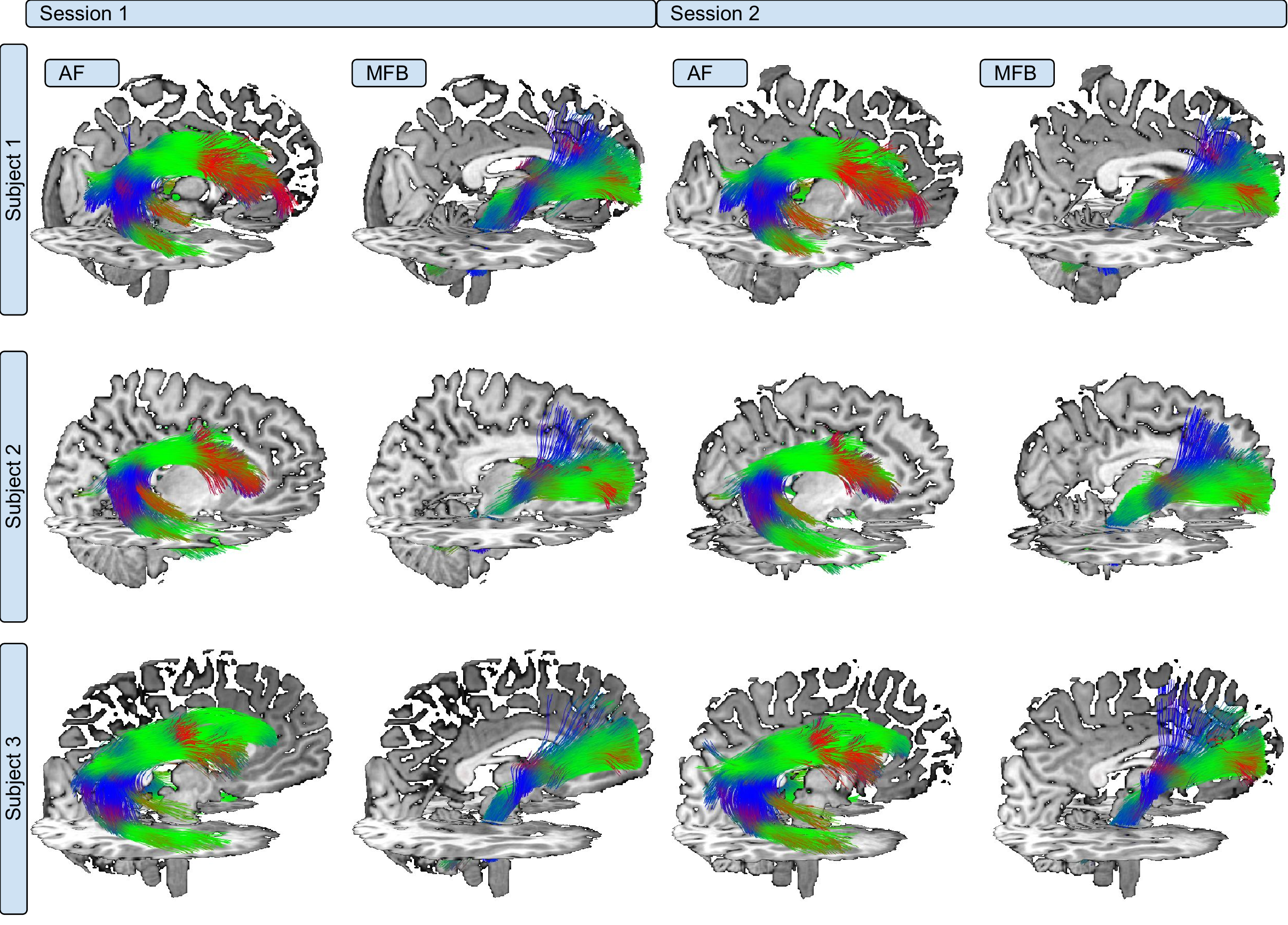}
    \end{center}
    \caption{Results for the two sessions acquired with the 
    TRIO protocol. The HAMLET machine was trained 
    on PRISMA data. For demonstration of the AF and MFB are selected 
    on three randomly chosen subjects. Visual similarity of the subjects 
    is apparent.}
    \label{fig:retest}
\end{figure*}

The TRIO dataset consists of 28 healthy volunteers scanned twice in 
two different sessions. We use this dataset to quantify the 
reproducibility of the obtained tract images, and second, to 
show that a trained machine does also work for protocols 
that are quite different compared to the one used for training.
Compared to the PRISMA dataset the TRIO protocol uses partial Fourier, 
has a different resolution, different echo time and different 
distortion corrections. The diffusion 
weighting schemes are approximately the same (see details above).
Figure \ref{fig:retest} shows the AF and MFB for three random 
subjects and the two different sessions. Both tracts are cleanly
detected and the appearance and shape of the tract is well 
preserved between the sessions. 

\begin{figure*}
    \begin{center}
    \includegraphics[width=14cm]{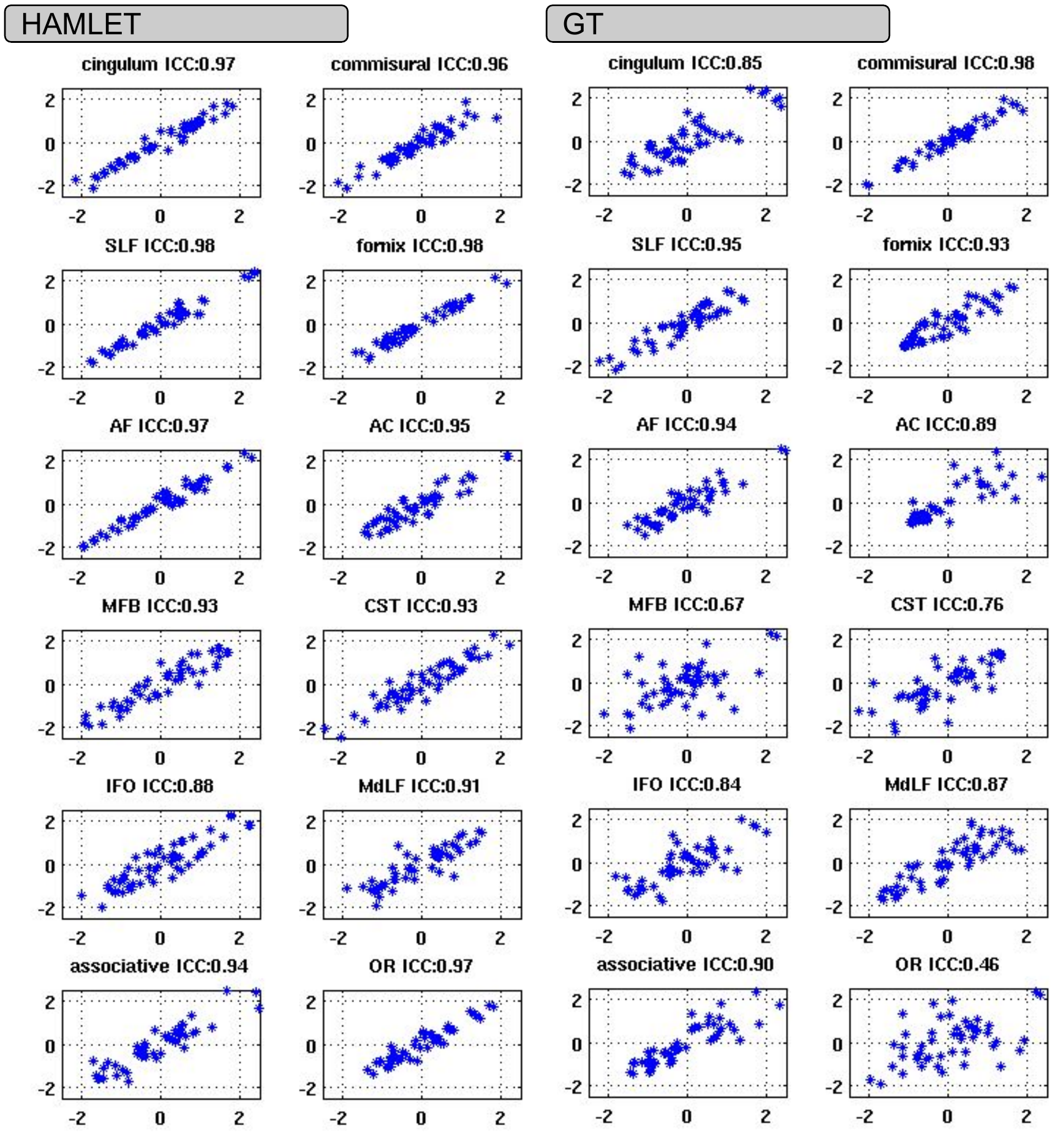}
    \end{center}
    \caption{Correlation plots between the two sessions for all 12 considered
    tracts for HAMLET and GT. The ICC for HAMLET a mostly above 0.9, while ICC 
    for GT drop for difficult tracts below an acceptable range.}
    \label{fig:retest_plots}
\end{figure*}

To measure the 'size' or 'volume' of the tract as a biomarker one 
can follow a simple strategy: if $\mv y(\mv r)$ is the tract
image, we just calculated the squared power $\int |\mv y(\mv r)|^2 d\mv r$
to measure an apparent tract volume. In Figure \ref{fig:retest_plots}
we show correlation plots between the sessions for all considered
tracts. For reference we show the streamline counts obtained from 
the automatic selection strategy based on GT. To quantify  
reproducibility intra-class correlation coefficients\footnote{
Used definition of ICC: $x^1_i$ and $x^2_i$ are values from 
first and second session, then ICC is defined by
$ICC = 1-\sum_i (x^1_i-x^2_i)^2 / \sum_{ij} (x^j_i - \bar x)^2$
} (ICC) are reported for HAMLET's apparent tract volumes and 
the streamline count obtained from GT.

%
%

\section{Discussion}
The detection and classification of structures in dMRI is challenging because of the low spatial resolution, the large image size and strong acquisition noise. The difficult, time consuming and error prone task of creating high quality training data makes it even more challenging to solve such problems in a machine learning driven manner. For a long time, the usage of machine learning techniques for solving classification tasks in dMRI data was mostly restricted to the classification of rather simple problems, like the classification into brain gray-matter and brain white-matter \cite{schnell2009fully}. With the contemporary remarkable improvements in machine learning, it is very recent that it became possible to perform complex classification tasks in dMRI, such as the localization of whole fiber tracts \cite{wasserthal2017direct,tractseg}.

In this context, we have presented HAMLET, a novel machine learning approach that tackles the challenging problem of dMRI-based fiber tracing. in contrast to the state-of-the-art, HAMLET provides both the tract location and tract directions. The combination of localization (tract position) and tract orientation is, in general, a difficult, memory demanding problem due to its high dimensionality in 3D. However, with the novel, rotation covariant training framework, HAMLET has been tailored to the learning of the concept of complicated fiber tracts with a small number of free parameters in short time.  

This is contrary to the trend of modern CNN architectures, like \cite{cciccek20163d}, which are very generic functional regressors with
millions of unknowns. To train a CNN on biomedical imagery usual requires a large number of training samples to let the network learn the underlying pattern statistics, and to learn the underlying concept of possible image transformations. An incomplete or incorrect representation of the underlying concepts is often the reason for over-fitting or poor generalization. A common method to regularize the training is the usage of image augmentation, like the on-the-fly generated deformations enable quick generalization. However, the optimization of such high-dimensional problems is frequently based on heuristics, and therefore it is currently impractical to prove whether a CNN has correctly learned the underlying transformation concepts or not. Further, training from scratch usually takes days. 

In fact, the deformations used for data augmentations
are usually diffeomorphisms and locally very close to simple rotations. 
So, it is quite nearby to require the predictor to be intrinsically 
robust, or even covariant against rotations. 
The strict covariance constraint can drastically decrease the parametric complexity of the whole regressor (not necessarily the computational
complexity). We followed the rather radical idea of full rotation and
parity covariance. In its current setting HAMLET needs about 1000
parameters to describe a tract, which is three orders of magnitudes less compared to CNNs used for 3D segmentations on comparable sized volumes (e.g. \cite{cciccek20163d,tractseg}). We found that already a few training examples without any augmentations are enough 
and a reasonable training session is accomplished within an hours.
If you turn it up-side-down, the space of possible tracts can be described within a 1000-dimensional space. We have also seen that the covariance makes the obtained apparent tract volumes very stable and an ideal candidate for a quantitative biomarker.

On the other hand, the problems of the approach have also been highlighted: responses are rather smooth and the performance depends on the structure to be learned. Of course, it is not surprising that certain structures can be defined better than others if only relative spatio-orientational information is available. It is also not very surprising that a machine with about 1000 parameters produces rather rough segmentations.
However, we also have seen that the simplicity of the machine makes it to generalize very well. Additionally, the 'ground truth' itself is a problem in general. Variabilities within the training set are not just caused by the 'true' natural variability, but also by errors during its generation.

Memory is also an issue, one application of HAMLET takes about 
4 GBs of memory, and learning is rather limited by memory than by speed, as long as no batch-learning is used. In the current setting the training of 20 subjects needs about 60 GBs of data concurrently held in memory.
However, improvements using batch-learning are nearby. 
%

\section{Conclusion}
This is a feasibility study to show that fully automatic, rotation covariant dMRI tract learning based on Spherical Tensor Algebra is possible.
The input of the machine is very close to the 
raw data and not very demanding with respect
to the acquisition protocol. 
The bundle anatomies follow the ones obtained
from conventional tractography and their volumes are stable across 
sessions, which makes the application as 
a quantitative biomarker possible. Instead of predicting scalar fields the prediction of rank-2 
tensor fields offers the opportunity to perform visually appealing streamline
tracking on the predicted tract images.
Applications to neuro-navigation are nearby.

For future work, it might be reasonable to drop partially the rotation covariance constraint as it makes the problem quite hard and rough normalizations can usually be determined in a reliable way. Taking higher order input might 
also be an option, however, this will limit the number of possible protocols. We only learned symmetric tracts; learning asymmetry is also possible, however, we found in preliminary experiments that it is probably not worth to directly learn asymmetric tracts, because it is hard to avoid contralateral false positives.
And actually it is not necessary that the machine solves the left/right problem for each label again 
and again. Labels could share the prediction of 
of learned presegmentations of left and right hemisphere. One could even go further and learn,
in an unsupervised way, rough, initial brain parcellations. The actual tract learning would 
just get these parcellations as additional input.
Also concentrating on certain parts by subdividing the brain could be a way to increase the resolution of the predictions.

\appendix
\section{Appendix}
\subsection{Product Selection} \label{sec:product}
There are a manifold of different ways to combine products and 
initial convolutions, however, not all of them can be robustly 
computed. For example, large initial smoothing allow higher order differentiations,
while tight kernels only allow for low order expansions.
In the following we explain the gritty details
how different smoothings and products are chosen.

Recall, the initial linear features are spherical tensor fields 
of order $J$ and obtained as follows
\[
\mv b_{J,L,j,g} = \PA{J}{\gmv \partial^L}{(g * \mv a^{j})},
\]
and are combined quadratically by 
\begin{eqnarray}
\mv f = \PA{2}{\gmv \partial^K }{(g'' * (\PA{\ell}{\mv b_{J,L,j,g}}{\mv b_{J',L',j',g'}}) )} \label{eq:quadfeat}
\end{eqnarray}
or linearly with the output of the previous layers
\begin{eqnarray}
\mv f = \PA{2}{\gmv \partial^K }
                {(g'' * (\PA{\ell}{\mv b_{J,L,j,g}}{\mv y_k}) )} \label{eq:linfeat}
\end{eqnarray}
Here $g,g'$ denote the radially symmetric initial smoothings. We work with Gaussians and Laplacians of Gaussians. Let's call this set 
$\mathcal{G}$. The set $\mathcal{G}$ depends on the level of 
the filter. The final smoothing $g''$ is a fixed Gaussian
with width of $1$ voxel. We denote a Gaussian of width $\sigma$ by $G_\sigma$
and a Laplacian of Gaussian by $G^\Delta_\sigma$

The free indices $j,j',L,L',J,J',K$ can vary according to the triangle 
inequality (or for higher levels $j,L,J,K,k$). To draw a cutoff we restricted $J$ and $J'$ to be smaller than a certain limit $\mathcal{J}$, while the other free indices can take all possible (parity covariant) values. 
For different smoothing kernels $g,g'$, it makes sense to adapt 
the cutoff, so $\mathcal{J}_g = \mathcal{J}$.
The cutoff is also depending on the filter level and decreases with resolution.

The following inequalities conclude the selection criteria
\begin{eqnarray}
	j,j' \in \{ 0,2 \} &\text{or}& k \in \{1,2,3\}\\
    g,g' &\in& \mathcal{G}\\
    g'' &=& G_{\sigma=1} \\
    J \leq  \mathcal{J}_g  &\text{and}&
    J \leq  \mathcal{J}_{g'} \\
    |j-L| \leq &\ell& \leq |j+L| \\
    |j'-L'| \leq &\ell& \leq |j'+L'| \\
    |\ell-K| \leq &2& \leq |\ell+K| \\
     \ell+J+J'+L &\text{is}& \text{\  even}
\end{eqnarray}
The used filter uses three layers, in Table XX we conclude the level
specific parameters. We additionally report $N$, the number of features 
quadratic in the signal (obtained from equation \eqref{eq:quadfeat}) 
and $N_s$, the number of features obtained by combining signal and 
the output from the previous layer (from equation \eqref{eq:linfeat}).
Note that, the latter have to be computed for each tract label 
individually, while the other do not depend on the previous layer's
output, and hence, are independent of the label to be learned.
\begin{center}
\begin{tabular}{|c|c|c|c|c|c|}
\hline
level & $g,g'$ & $\mathcal{J}$ & $N$ & $N_s$ & $\gamma^s N$ \\ \hline\hline
\multirow{ 3}{*}{s=-2}    & $G_1$  & 5 & \multirow{ 3}{*}{639} 
										 & \multirow{ 3}{*}{0} & \multirow{ 3}{*}{260MB}\\
                          & $G^\Delta_{1.5}$  & 5 &&& \\
                          & $G^\Delta_{2}$  & 5 &&& \\ \hline
\multirow{ 3}{*}{s=-1}    & $G_1$  & 3 & \multirow{ 3}{*}{219}
										 & \multirow{ 3}{*}{32}& \multirow{ 3}{*}{690MB}\\
                          & $G^\Delta_{1.5}$  & 3 &&& \\
                          & $G^\Delta_{2}$  & 3 &&& \\ \hline
\multirow{ 2}{*}{s=-1}    & $G_1$  & 3 & \multirow{ 2}{*}{102}
										 & \multirow{ 2}{*}{24}& \multirow{ 2}{*}{2500MB}\\
                          & $G^\Delta_{2}$  & 3 &&& \\ \hline
\end{tabular}
\end{center}
  
\subsection{Details on Processing and Learning} \label{sec:learning}
In the following we will give a more detailed description of the actual processing and 
learning steps. 

As computational load depends strongly on the size of the processed
volume, we used MNI coordinates to automatically generate a minimal bounding box.
To avoid boundary effects and fold overs due to the circularity of FFT-based convolutions an
additional margin had to be respected (of about 3 voxels).
Each volume is resampled to a resolution 1.5mm isotropic and windowed by a smooth windowing 
function.
If no MNI coordinates are available,
alternative brain extraction tools could also be used (e.g. by the use of FSL's BET). 
Depending on the scale, we apply a Gaussian smooth before we resample the raw data. If 
$s$ is the scale, we smooth by $0.75-\gamma^s\sigma$.
The resampling of the label images is done accordingly on each scale. 

The learning itself is based on a simple step-wise linear training.
Let  $\mv F_m$ be five matrices representing $m$th spherical component 
of the collected features $\mv f \in \Complex^5$ from all subjects. 
That is, the matrix $\mv F_m$ is of size $(N) \times (\sum_i M_i)$ where $M_i$ is the number of 
voxels in the $i$th subject and $N$ the number of features. 

And further let $\mv Y_m$ the label vectors, each of size $\mv \sum_i M_i$. Then, we optimize the following
linear objective
\[
J(\gmv \alpha) =  \sum_{m=-2}^{m=2} |\mv Y_m -  \mv F_m \gmv \alpha|^2 + \lambda |\gmv \alpha|^2
\]
where $\gmv \alpha$ are the parameters to be learned. For the lowest scale
$\gmv \alpha = \{ \alpha_{jII'} \}$,
or $\gmv \alpha = \{\kappa, \alpha_{jII'} , \alpha_{jIk}  \}$ for the higher scales (see equation \eqref{eq:feats}).

Due to the high dynamic variance the features 
are normalized before inversion by their power 
(the square-root of the diagonal of $\sum_m \mv F_m^\T \mv F_m$). So, also the regularization have
to be seen in this respect, relative to the power of the feature. 
The conditioning of the problem usually gets better with higher scales,
because dimensionality decreases and the number of examples increases. 
We used the same regularization of $\lambda=0.001$ for all scales, which might be not optimal, but simple. We found that differences 
in performance are not very prominent as long as $\lambda$ is not too low.
Using the pseudoinverse is also an option, but would lengthen the computation time.

\subsection{Tract Selection} \label{sec:tract}
Supposing MNI coordinates are given: a tract, namely a set of streamlines,
is determined by a sequence of MNI coordinates
$\mv x_i\in \Real^3$ and tolerances $t_i$ with $i<n$. A arc-length parametrized streamline $\mv r(t)$ belongs to a tract, if 
\[
|\mv r(t_i) - \mv x_i| < t_i \ \forall i: \ 0 \leq i < n
\]
where $t_i = \sum_{j<i} |\mv x_j - \mv x_{j+1}|$. Here the list of 12 tracts used in 
this study:

\begin{center}
\begin{tabular}{|l|c|c|c|c|}
\hline
name & $x$ & $y$ & $z$ & $t$ \\ \hline\hline
{MFB}    & see \cite{coenen2018anatomy}  &&&\\ \hline
\multirow{ 3}{*}{cingulum}    & -30  & -22 & -22 & 14 \\
                              & -11 & -47 & 17 & 5 \\
                              & -17  & 39 & 17 & 12 \\ \hline
\multirow{ 3}{*}{Sup. Long. Fasc. (SLF)}    & -45  & 30 & 22 & 8 \\
					    & -39  & -24 & 36 & 6 \\
					    & -45  & -71 & 40 & 12 \\ \hline
\multirow{ 3}{*}{Arcuate Fascicle (AF)}     & -48  & 22  & 10 & 10 \\
					    & -40  & -10 & 22 & 5 \\
					    & -36  & -42 & 12 & 4 \\
					    & -55  & -12 & -13 & 7 \\ \hline
\multirow{ 3}{*}{IFOF }                     & -27  & 52 & 3 & 10 \\
					    & -33  & -16 & -11 & 4 \\
					    & -23  & -84 & -8 & 12 \\ \hline
\multirow{ 3}{*}{MdLF }                     & -57  & -10 & 4 & 5 \\
					    & -34  & -37 & 20 & 4 \\
					    & -25  & -57 & 49 & 8 \\ \hline
\multirow{ 3}{*}{anterior commisure (AC)}   & -30  & -78 & -25 & 15 \\
					    &  0  & 2 & -6 & 5 \\
					    & 30  & -78 & -25 & 15 \\ \hline
\multirow{ 3}{*}{commissural central}   & 40  & 0 & 51 & 13 \\
					&  0  & 0 & 24 & 2 \\
					& -40  & 0 & 51 & 13 \\ \hline
\multirow{ 3}{*}{optic radiation}       & -6  & -31 & -9 & 3 \\
					&  -32  & -31 & -3 & 3 \\
					&  -32  & -56 & 4 & 6 \\
					& -23  & -90 & 9 & 9 \\ \hline
\multirow{ 3}{*}{fornix}       		& -19  & -4 & -15 & 6 \\
					& 0    & -11 & 17 & 4 \\
					& -20  & -32 & 6 & 5 \\
					& -30  & -17 & -20 & 6 \\ \hline
\multirow{ 3}{*}{CST}       		& 33  & -12 & -59 & 17 \\
					& 19    & -14 & -5 & 2 \\
					& 7  & -27 & -41 & 5 \\ \hline
\multirow{ 3}{*}{FAT}       		& -55  & 12 & 13 & 4 \\
					& -30  & 5 & 24 & 4 \\
					& -25  & 7 & 58 & 10 \\ \hline
\end{tabular}
\end{center}
All selections where taken from a global tractography with 5-fold accumulated tracts.

\bibliographystyle{plain}      

\bibliography{literatureMRI,mylit}

\begin{thebibliography}{10}

\bibitem{andersson2003correct}
Jesper~LR Andersson, Stefan Skare, and John Ashburner.
\newblock How to correct susceptibility distortions in spin-echo echo-planar
  images: application to diffusion tensor imaging.
\newblock {\em Neuroimage}, 20(2):870--888, 2003.

\bibitem{cciccek20163d}
{\"O}zg{\"u}n {\c{C}}i{\c{c}}ek, Ahmed Abdulkadir, Soeren~S Lienkamp, Thomas
  Brox, and Olaf Ronneberger.
\newblock 3d u-net: learning dense volumetric segmentation from sparse
  annotation.
\newblock In {\em International Conference on Medical Image Computing and
  Computer-Assisted Intervention}, pages 424--432. Springer, 2016.

\bibitem{coenen2018anatomy}
Volker~Arnd Coenen, Lena~Valerie Schumacher, Christoph Kaller, Thomas~Eduard
  Schlaepfer, Peter~Christoph Reinacher, Karl Egger, Horst Urbach, and Marco
  Reisert.
\newblock The anatomy of the human medial forebrain bundle: Ventral tegmental
  area connections to reward-associated subcortical and frontal lobe regions.
\newblock {\em NeuroImage: Clinical}, 18:770--783, 2018.

\bibitem{hagler2009automated}
Donald~J Hagler, Mazyar~E Ahmadi, Joshua Kuperman, Dominic Holland, Carrie~R
  McDonald, Eric Halgren, and Anders~M Dale.
\newblock Automated white-matter tractography using a probabilistic diffusion
  tensor atlas: Application to temporal lobe epilepsy.
\newblock {\em Human brain mapping}, 30(5):1535--1547, 2009.

\bibitem{jorgens2018learning}
Daniel J{\"o}rgens, {\"O}rjan Smedby, and Rodrigo Moreno.
\newblock Learning a single step of streamline tractography based on neural
  networks.
\newblock In {\em Computational Diffusion MRI}, pages 103--116. Springer, 2018.

\bibitem{kellner2016gibbs}
Elias Kellner, Bibek Dhital, Valerij~G Kiselev, and Marco Reisert.
\newblock Gibbs-ringing artifact removal based on local subvoxel-shifts.
\newblock {\em Magnetic resonance in medicine}, 76(5):1574--1581, 2016.

\bibitem{koppers2017reconstruction}
Simon Koppers, Matthias Friedrichs, and Dorit Merhof.
\newblock Reconstruction of diffusion anisotropies using 3d deep convolutional
  neural networks in diffusion imaging.
\newblock In {\em Modeling, Analysis, and Visualization of Anisotropy}, pages
  393--404. Springer, 2017.

\bibitem{koppers2016direct}
Simon Koppers and Dorit Merhof.
\newblock Direct estimation of fiber orientations using deep learning in
  diffusion imaging.
\newblock In {\em International Workshop on Machine Learning in Medical
  Imaging}, pages 53--60. Springer, 2016.

\bibitem{maier2017challenge}
Klaus~H Maier-Hein, Peter~F Neher, Jean-Christophe Houde, Marc-Alexandre
  C{\^o}t{\'e}, Eleftherios Garyfallidis, Jidan Zhong, Maxime Chamberland,
  Fang-Cheng Yeh, Ying-Chia Lin, Qing Ji, et~al.
\newblock The challenge of mapping the human connectome based on diffusion
  tractography.
\newblock {\em Nature communications}, 8(1):1349, 2017.

\bibitem{neher2015machine}
Peter~F Neher, Michael G{\"o}tz, Tobias Norajitra, Christian Weber, and Klaus~H
  Maier-Hein.
\newblock A machine learning based approach to fiber tractography using
  classifier voting.
\newblock In {\em International Conference on Medical Image Computing and
  Computer-Assisted Intervention}, pages 45--52. Springer, 2015.

\bibitem{poulin2017learn}
Philippe Poulin, Marc-Alexandre Cote, Jean-Christophe Houde, Laurent Petit,
  Peter~F Neher, Klaus~H Maier-Hein, Hugo Larochelle, and Maxime Descoteaux.
\newblock Learn to track: Deep learning for tractography.
\newblock In {\em International Conference on Medical Image Computing and
  Computer-Assisted Intervention}, pages 540--547. Springer, 2017.

\bibitem{reisert2010}
M.~Reisert, I.~Mader, C.~Anastasopoulos, M.~Weigel, S.~Schnell, and V.~Kiselev.
\newblock Global fiber reconstruction becomes practical.
\newblock {\em Neuroimage}, 54(2):955--62, 2011.

\bibitem{reisert2017disentangling}
Marco Reisert, Elias Kellner, Bibek Dhital, Juergen Hennig, and Valerij~G
  Kiselev.
\newblock Disentangling micro from mesostructure by diffusion mri: A bayesian
  approach.
\newblock {\em NeuroImage}, 147:964--975, 2017.

\bibitem{schnell2009fully}
Susanne Schnell, Dorothee Saur, BW~Kreher, J{\"u}rgen Hennig, H~Burkhardt, and
  Valerij~G Kiselev.
\newblock Fully automated classification of hardi in vivo data using a support
  vector machine.
\newblock {\em NeuroImage}, 46(3):642--651, 2009.

\bibitem{skibbespherical}
Henrik Skibbe and Marco Reisert.
\newblock Spherical tensor algebra: A toolkit for 3d image processing.
\newblock {\em Journal of Mathematical Imaging and Vision}, pages 1--33.

\bibitem{skibbe2011dense}
Henrik Skibbe and Marco Reisert.
\newblock Dense rotation invariant brain pyramids for automated human brain
  parcellation.
\newblock In {\em Proceedings of the Workshop on Emerging Technologies for
  Medical Diagnosis and Therapy (Informatik???11)}, 2011.

\bibitem{skibbe2011gaussian}
Henrik Skibbe, Marco Reisert, and Hans Burkhardt.
\newblock Gaussian neighborhood descriptors for brain segmentation.
\newblock In {\em MVA}, pages 35--38, 2011.

\bibitem{veraart2016denoising}
Jelle Veraart, Dmitry~S Novikov, Daan Christiaens, Benjamin Ades-Aron, Jan
  Sijbers, and Els Fieremans.
\newblock Denoising of diffusion mri using random matrix theory.
\newblock {\em Neuroimage}, 142:394--406, 2016.

\bibitem{tractseg}
Jakob Wasserthal, Peter Neher, and Klaus~H Maier-Hein.
\newblock Tractseg-fast and accurate white matter tract segmentation.
\newblock {\em arXiv preprint arXiv:1805.07103}, 2018.

\bibitem{wasserthal2017direct}
Jakob Wasserthal, Peter~F Neher, Fabian Isensee, and Klaus~H Maier-Hein.
\newblock Direct white matter bundle segmentation using stacked u-nets.
\newblock {\em arXiv preprint arXiv:1703.02036}, 2017.

\end{thebibliography}

%

\end{document}